# The Generator-Eraser Paradox: Community Guidelines for Responsible LLM-Assisted Dialect Resource Creation

**Wajdi Zaghouani**
Northwestern University in Qatar
wajdi.zaghouani@northwestern.edu

**Abstract**

Dialect resources occupy a unique position at the intersection of scientific description, cultural preservation, and computational infrastructure. Large language models offer powerful capabilities for accelerating dialect resource development through retrieval-grounded drafting, corpus navigation, metadata enrichment, and annotation workflow support. However, the same systems pose substantial risks: they can contribute to dialect erasure by privileging prestige varieties, homogenizing orthography, and enabling synthetic feedback loops that reduce linguistic diversity over time. These risks are particularly acute for language varieties characterized by diglossia, limited written standardization, or marginalized speaker communities. This paper makes three contributions. First, we integrate insights from variationist sociolinguistics and corpus linguistics to formalize the **generator-eraser paradox** as a theoretical framework for understanding the dual nature of LLM-assisted dialect work. Second, we derive **12 community guidelines** that operationalize this framework into implementable design requirements for dialect resource creation and documentation. Third, we provide an **in-depth case study of Arabic dialects**, including a structured comparison of widely used resources, to demonstrate how these guidelines address language-specific challenges including diglossia, orthographic variability, and community governance. The contribution is conceptual and operational rather than experimental, with the goal of enabling dialect communities and resource builders across languages to adopt LLMs without sacrificing authenticity, variation, or sovereignty.

## 1. Introduction

Language variation is not noise; it is the systematic mapping of linguistic forms to social meanings [Labov, 1972, Eckert, 2012]. Dialects, regional varieties, and sociolects encode community identity, geographic affiliation, social positioning, and cultural knowledge in ways that standard varieties cannot capture. For computational linguistics, dialect resources serve multiple functions: they enable training of dialect-aware NLP systems, provide evaluation benchmarks for measuring system fairness, support language documentation and preservation efforts, and create infrastructure for communities to engage with technology on their own terms.

Yet dialect resources remain chronically underdeveloped relative to standardized varieties. The resource gap documented by Joshi et al. [2020] applies not only across languages but within languages, where prestige varieties dominate corpora, evaluation sets, and model training data. This imbalance has measurable consequences. Automatic speech recognition systems exhibit higher error rates for speakers of African American English [Koenecke et al., 2020]. Hate speech detection systems disproportionately flag dialect features as toxic [Sap et al., 2019]. Part-of-speech taggers trained on standard varieties degrade when applied to dialect text [Hovy and Søgaard, 2015, Blodgett et al., 2016]. The common thread is that systems optimized for prestige varieties treat dialect features as deviations rather than as systematic linguistic patterns.

Large language models introduce both opportunity and risk into this landscape. On one hand, LLMs can accelerate resource creation: they can assist in drafting lexicographic entries from retrieved attestations, propose candidate annotations for human confirmation, generate pedagogical examples grounded in dialect corpora, and help navigate large collections of unstructured dialect data. On the other hand, LLMs are trained on corpora that reflect and amplify existing resource imbalances. Their statistical priors favor high-frequency forms, which in language contexts typically means prestige varieties. When deployed in resource creation workflows, they risk producing outputs that drift toward dominant norms, homogenize orthographic variation, and introduce synthetic material that, if recursively incorporated into future training, can narrow linguistic distributions and eliminate precisely the tail phenomena that constitute dialect distinctiveness [Shumailov et al., 2024].

This paper addresses a gap in the literature: the absence of a systematic framework for governing LLM involvement in dialect resource work. Our contributions are threefold:

1. We develop the **generator-eraser paradox** as a theoretical framework that synthesizes variationist sociolinguistics, corpus linguistics, and recent findings on model collapse to articulate the dual nature of LLM assistance in dialect contexts.

2. We derive **12 community guidelines** from this framework, written as implementable design requirements with explicit checks and language-specific considerations.
3. We provide an **in-depth case study of Arabic dialects**, demonstrating how the guidelines address challenges of diglossia, orthographic variability, and multi-stakeholder governance that generalize to other diglossic and pluricentric language situations.

The paper is deliberately non-experimental. We do not report model performance metrics, and we avoid unverifiable claims about system behavior. Instead, we translate stable findings from sociolinguistics, corpus linguistics, and AI governance into operational requirements for responsible dialect resource development.

# 2. Background

## 2.1. Dialect Variation as Structured Diversity

Three waves of variationist sociolinguistics have established that dialect variation is systematic rather than random [Eckert, 2012]. The first wave, pioneered by Labov [1972], documented correlations between linguistic variables and social categories such as class, age, and gender. The second wave examined local communities and social networks, revealing how dialect features index local identity and group membership. The third wave reframed variation as a resource for constructing social meaning, emphasizing speaker agency in deploying dialect features for communicative purposes [Chambers, 2013].

For resource creation, these findings imply that dialect corpora are not merely collections of sentences; they are archives of variation. The resource object includes linguistic forms and their frequencies, the contexts and registers in which forms appear, metadata about speakers and communities where ethically available, and the orthographic practices and writing conventions that speakers employ. Any processing pipeline that alters these dimensions without traceability risks severing the link between form and social meaning.

## 2.2. Diglossia and Prestige Hierarchies

Many language situations involve diglossia: a stable relationship between a high variety associated with formal contexts and low varieties associated with everyday speech [Ferguson, 1959]. Arabic exemplifies classical diglossia, with Modern Standard Arabic occupying institutional domains while regional dialects dominate spoken interaction [Holes, 2004, Versteegh, 2014]. However, diglossia is not unique to Arabic. Similar configurations appear in Swiss German, Haitian Creole, Tamil, and many other language situations where written standards diverge from spoken vernaculars.

Diglossia creates a prestige gradient that shapes both human behavior and computational systems. When models or preprocessing pipelines attempt to normalize text, they tend to normalize toward the institutionally privileged variety. In resource work, such normalization can be an explicit design choice, such as adopting a conventional orthography, or an implicit side effect of preprocessing steps such as spell-checking, tokenization, or automatic correction [Eisenstein, 2013].

## 2.3. Corpus Representativeness and Provenance

Corpus linguistics distinguishes size from representativeness. A corpus must be designed or sampled to reflect intended domains, registers, and speaker groups, and it must include metadata enabling audit and interpretation [Biber, 1993, McEnery and Hardie, 2012]. Dialect corpora are frequently convenience samples from social media platforms with strong demographic and topical skews. These skews may favor urban over rural speakers, younger over older generations, particular countries or regions with higher platform penetration, and specific genres such as political commentary or entertainment.

Provenance matters because dialect variation is socially indexed. Without provenance, a resource cannot be reliably interpreted as a proxy for a dialect community; it becomes a proxy for platform use. Documentation frameworks such as datasheets for datasets [Gebru et al., 2021] and data statements [Bender and Friedman, 2018] provide templates for recording this information, but adoption remains uneven in dialect resource work.

## 2.4. LLM Risks: Representational Harm and Synthetic Recursion

Two distinct bodies of research document LLM risks relevant to dialect resources.

First, work on representational harms shows that language models reproduce and amplify biases present in training data [Blodgett et al., 2020, Bender et al., 2021]. In dialect settings, this includes bias toward prestige norms, stereotyped associations between dialect features and social categories, and performance degradation on dialect text that cascades through downstream applications [Jurgens et al., 2017, Tan et al., 2020].

Second, research on model collapse demonstrates that training on recursively generated data can eliminate low-probability events, leading to progressively impoverished distributions [Shumailov et al., 2024]. For dialect resources, low-probability events often correspond to precisely what makes dialects distinctive: minority lexical items, region-specific constructions, and socially marked forms that occur infrequently in any given corpus but carry substantial social meaning.

### 2.5. Existing Dialect Resources and Shared Tasks

The NLP community has developed substantial infrastructure for dialect processing. The VarDial workshop series and associated shared tasks have produced benchmarks for dialect identification across multiple language families [Zampieri et al., 2017]. For Arabic specifically, resources include the Arabic Online Commentary Dataset [Zaidan and Callison-Burch, 2011], the MADAR multi-dialect corpus and lexicon [Bouamor et al., 2018], the Gumar Gulf Arabic corpus [Khalifa et al., 2016], and shared tasks such as NADI [Abdul-Mageed et al., 2021]. Multi-dialect resources have also been developed for tasks such as author profiling [Zaghouani and Charfi, 2018], fine-grained dialect annotation [Charfi et al., 2019], and hate speech detection across Arabic varieties [Charfi et al., 2024a, Zaghouani et al., 2024]. For African American English, TwitterAAE provides a large-scale resource with demographic annotations [Blodgett et al., 2016]. Earlier surveys of freely available Arabic corpora [Zaghouani, 2014] have documented the landscape of accessible resources, revealing both the breadth and the gaps in existing coverage.

These resources demonstrate mature methodological practices but also reveal fragmentation. Different datasets adopt different orthographies, annotation frameworks, and access conditions. This fragmentation creates barriers to scalable dialect resource development and cross-resource interoperability.

## 3. The Generator-Eraser Paradox

We propose the **generator-eraser paradox** as a framework for understanding the dual nature of LLM involvement in dialect resource work. The paradox captures a fundamental tension: the same properties that make LLMs useful for resource creation also make them dangerous for dialect preservation.

### 3.1. The Dual Nature of LLM Assistance

LLMs can accelerate dialect resource development through several mechanisms. *Retrieval-grounded drafting* allows LLMs to draft lexicographic entries, example sentences, or educational materials grounded in retrieved dialect attestations; retrieval-augmented generation [Lewis et al., 2020] reduces hallucination and anchors outputs in authentic dialect data. *Corpus navigation and clustering* helps resource builders explore large unstructured collections, identify patterns, and cluster similar items for batch annotation. *Metadata enrichment* enables LLMs to propose metadata such as topic labels, register classifications, or dialect sub-variety tags for human confirmation. *Annotation workflow support* allows LLMs to generate candidate annotations that human experts review, revise, and approve, reducing annotation time while maintaining human authority over final decisions.

However, the same LLMs can erode dialect distinctiveness through countervailing mechanisms. *Prior-driven normalization* occurs because LLMs trained predominantly on prestige varieties have statistical priors that favor standard forms; without explicit constraints, generated outputs drift toward these priors. *Orthographic homogenization* reduces the variability that reflects phonological differences, community conventions, or the absence of written standardization. *Synthetic recursion* occurs when LLM-generated content enters training pipelines for future models or becomes part of dialect corpora without adequate labeling; model collapse research shows this effect is particularly severe for low-frequency events [Shumailov et al., 2024]. *Prestige drift*, which we define as the systematic substitution of dialect-specific markers (phonological, morphological, lexical, or syntactic) with high-variety equivalents over iterative processing, may be imperceptible in any single interaction but cumulative over project lifecycles.

### 3.2. Amplifying Factors

Several factors amplify the paradox in particular language situations. In *diglossic* settings, the high variety dominates written corpora and institutional language, creating severe training data imbalances that LLMs inherit and reproduce unless explicitly constrained. Dialects with *limited standardization* exhibit higher orthographic variability and less consistent lexicographic documentation; LLMs may treat this variability as error rather than as authentic variation. *Platform sampling* from social media introduces demographic, topical, and temporal skews that LLMs trained on such data may amplify. *Community power differentials* mean that dialect communities often lack institutional power to shape how their language is documented and processed; without participatory governance, external resource builders may make decisions that do not reflect community priorities.

# 4. Design Requirements

This section converts the theoretical framework into explicit design requirements that the proposed guidelines must address.

**Preserving Distributional Properties.** Dialect resources must preserve the distributional properties of authentic language use, including the frequency and context of variant forms, the presence of low-frequency items, and the orthographic diversity of written representations.

**Layered Orthographic Representation.** Orthographic normalization should be implemented as a reversible transformation with documented mappings, not as a destructive rewrite.

**Transparent Provenance.** All resource artifacts should include clear provenance information distinguishing human-produced content from LLM-assisted content.

**Human Authority.** Human experts, particularly native speakers and community members, must retain authority over final resource decisions.

**Community Governance.** Dialect communities should participate in scope setting, annotation framework design, release decisions, and harm review [Costanza-Chock, 2020, Bird, 2020].

**Recursion Avoidance.** Synthetic data must be transparently labeled and must not enter training pipelines or corpus additions without explicit evaluation of diversity impacts.

# 5. The 12 Community Guidelines

Each guideline is written as an implementable requirement with rationale, checks, and considerations for language-specific adaptation.

## 5.1. Guideline 1: Real Data Primacy

**Requirement.** Human-produced dialect data remains the primary substrate for all resource artifacts. Synthetic text can only serve as an explicitly labeled augmentation layer, never as a replacement for authentic data.

**Rationale.** Model behavior follows training distributions and resource composition [Bender et al., 2021]. Synthetic recursion can eliminate tail events that constitute dialect distinctiveness [Shumailov et al., 2024].

**Checks.** Maintain a datasheet documenting collection methods, coverage, and limitations [Gebru et al., 2021]. Store synthetic content separately with full provenance metadata including prompts, models, and generation parameters.

**Adaptation.** For languages with minimal written dialect tradition, spoken transcripts may constitute primary data, with transcription conventions documented.

## 5.2. Guideline 2: Retrieval Before Generation

**Requirement.** For resource drafting tasks, prefer retrieval-augmented generation over unconstrained prompting. Outputs should cite or derive from authenticated dialect attestations.

**Rationale.** Retrieval-augmented generation conditions outputs on evidence rather than priors [Lewis et al., 2020], reducing drift toward prestige norms.

**Checks.** Log retrieval sources for generated content. Implement verification that generated forms appear in the retrieval set. Monitor retrieval quality through: (1) coverage metrics measuring what proportion of dialect marker types in outputs are attested in retrieved sources; (2) recall for low-frequency markers that are most vulnerable to prestige substitution; and (3) source diversity ensuring outputs draw from multiple attestations rather than over-relying on single sources.

**Adaptation.** For under-documented dialects, retrieval sets may be small; in such cases, additional human validation compensates for limited grounding. When retrieval fails to return sufficient dialect-specific attestations, flag outputs for mandatory expert review rather than allowing fallback to unconstrained generation.

## 5.3. Guideline 3: Orthographic Preservation

**Requirement.** Preserve raw orthographic forms as the archival layer. If normalization is required for processing, implement it as a reversible transformation with documented mappings.

**Rationale.** Orthography encodes phonological, social, and identity information. Conventional orthography frameworks such as CODA for Arabic [Habash et al., 2012] provide principled approaches to reversible normalization.

**Checks.** Maintain three layers: raw orthography, normalized orthography, and the mapping between them. Document what information is lost or ambiguous in normalization. Mappings should specify: (1) the normalization rule applied; (2) whether the mapping is one-to-one (fully reversible) or many-to-one (lossy); and (3) the linguistic rationale. For example, a Gulf Arabic raw form *chinna* normalized to CODA *kinna* would be documented as: `raw: chinna → norm: kinna | rule: ch→k | type: lossy (ch/k merger) | note: ch reflects /tš/ pronunciation`.

**Adaptation.** For languages with Latin script, normalization may address capitalization and diacritics; for non-Latin scripts, language-specific conventions apply.

### 5.4. Guideline 4: Dialect-Aware Prompting

**Requirement.** Prompts must explicitly specify dialect identity and instruct against implicit translation to prestige varieties. Prompts should request uncertainty marking rather than confident guessing.

**Rationale.** LLMs default to dominant norms under ambiguity. Explicit constraints can reduce prestige drift.

**Checks.** Include dialect specification in all prompts used for resource creation. Log prompts with outputs for audit.

**Adaptation.** Dialect names and boundaries vary by language; use community-preferred terminology.

### 5.5. Guideline 5: Human Validation Gate

**Requirement.** No LLM-assisted resource artifact is released without review by native speakers and domain experts. Review covers orthographic authenticity, semantic accuracy, pragmatic appropriateness, and cultural acceptability.

**Rationale.** Representational harms are systematic and unevenly distributed [Blodgett et al., 2020]. Only community-grounded review can detect subtle errors that automated metrics miss.

**Checks.** Define review rubrics. Record reviewer judgments and disagreements as signals of variation rather than errors to resolve.

**Adaptation.** Reviewer selection should prioritize diversity across dialect sub-varieties, speaker demographics, and expertise types.

### 5.6. Guideline 6: Synthetic Data Governance

**Requirement.** Synthetic data must be transparently labeled in all resource artifacts and must not be incorporated into recursive training pipelines without controlled evaluation of diversity impacts.

**Rationale.** Recursive training on synthetic data causes model collapse and diversity loss [Shumailov et al., 2024].

**Checks.** Implement content-level provenance using structured metadata fields (e.g., in JSON/JSONL schemas: `is_synthetic: boolean`, `generation_model: string`, `generation_date: timestamp`, `prompt_hash: string`). Maintain separate partitions for human and synthetic content. For large-scale pipelines, consider dataset-level synthetic ratio caps (e.g., ≤20% synthetic content) with periodic contamination audits.

**Adaptation.** For shared tasks, clarify whether synthetic augmentation is permitted and require participants to report synthetic content ratios and generation methods.

### 5.7. Guideline 7: Diversity Monitoring

**Requirement.** Track whether resource releases retain dialect markers and do not exhibit drift toward prestige varieties or dominant regional norms.

**Rationale.** Representativeness and balance require monitoring, not assumption [Biber, 1993, McEnery and Hardie, 2012].

**Checks.** Recommended approaches include: (1) compiling dialect marker inventories (diagnostic lexemes, morphological patterns, phonological spellings) validated by community experts; (2) computing marker frequency distributions across resource versions; (3) measuring distributional divergence using Jensen-Shannon divergence or similar metrics between current and baseline versions; (4) tracking type-token ratios for dialect-specific vocabulary; and (5) applying mixed-effects models controlling for topic and register when comparing releases. Flag statistically significant drift for manual review.

**Adaptation.** Marker selection should reflect language-specific features that distinguish dialects; community input is essential for validating marker inventories and setting acceptable drift thresholds.

### 5.8. Guideline 8: Participatory Governance

**Requirement.** Dialect communities participate in defining resource scope, annotation frameworks, release conditions, and harm review processes.

**Rationale.** Participatory design supports community self-determination and reduces extractive dynamics [Bird, 2020, Costanza-Chock, 2020].

**Checks.** Document community consultation processes. Include community representatives in project governance.

**Adaptation.** Community structure varies; some dialects have organized advocacy groups, others require identifying representative speakers through sociolinguistic fieldwork.

### 5.9. Guideline 9: Data Sovereignty and Access Control

**Requirement.** Clarify ownership and access rights for all resource components. Adopt tiered access when community safety, privacy, or cultural sensitivity requires restriction.

**Rationale.** Some dialect communities face political risk, discrimination, or cultural appropriation concerns. This guideline aligns with the CARE Principles for Indigenous Data Governance [Carroll et al., 2020].

**Checks.** Specify licensing terms. Document rationale for access restrictions. Consider benefit-sharing arrangements that return value to contributing communities.

**Adaptation.** Legal frameworks vary by jurisdiction; consult local expertise on data protection and indigenous language rights where applicable [Rainie et al., 2019].

### 5.10. Guideline 10: Pipeline Documentation

**Requirement.** Publish a pipeline card documenting all preprocessing steps, orthography handling, prompting strategies, retrieval configurations, model versions, and validation procedures.

**Rationale.** Transparency enables audit, reproduction, and identification of potential harm sources [Gebru et al., 2021, Mitchell et al., 2019].

**Checks.** Use structured documentation templates. Version control pipeline configurations.

**Adaptation.** Documentation should be accessible to both technical and community audiences; plain language summaries complement technical specifications.

### 5.11. Guideline 11: Prestige Drift Audits

**Requirement.** Conduct periodic audits to detect systematic replacement of dialect-specific forms with prestige variety equivalents.

**Rationale.** Drift may be imperceptible in individual outputs but cumulative over project lifecycles.

**Checks.** Audit procedures should include: (1) random sampling of outputs for manual expert review against dialect authenticity criteria; (2) automated comparison of dialect marker frequencies against baseline corpora using the metrics from Guideline 7; (3) targeted review of high-risk categories (e.g., function words, verbal morphology, discourse markers) where prestige substitution is most common; and (4) documentation of detected drift with root cause analysis (prompt design, retrieval gaps, or model limitations). Establish correction workflows for identified issues before release.

**Adaptation.** Audit frequency should reflect project pace; high-throughput workflows require more frequent checks, potentially automated continuous monitoring with periodic manual validation.

### 5.12. Guideline 12: Long-Term Stewardship

**Requirement.** Treat dialect resources as living infrastructure with periodic review, correction mechanisms, and versioned releases.

**Rationale.** Languages evolve, communities' needs change, and resource harms may emerge over time.

**Checks.** Establish maintenance schedules. Provide error reporting channels. Document version histories with changelogs.

**Adaptation.** Sustainability planning should consider institutional support, community capacity, and succession planning for key personnel.

## 6. Case Study: Arabic Dialects

Arabic provides a compelling case study because it exhibits the full range of challenges the guidelines address: classical diglossia, substantial orthographic variability, multiple regional varieties with distinct features, and complex political and cultural dynamics. This section demonstrates how the guidelines apply to Arabic while extracting generalizable lessons.

### 6.1. Diglossia, Orthography, and the Arabic Script System

Arabic diglossia involves Modern Standard Arabic (MSA) as the high variety used in formal writing, education, and media, alongside regional dialects that serve as primary spoken languages across the Arab world [Ferguson, 1959, Holes, 2004]. Subsequent scholarship emphasizes a continuum of registers and mixed styles, including extensive code-switching between MSA and dialect elements [Badawi, 1973]. For resource creation, this continuum means that dialect corpora frequently contain MSA elements, and annotation frameworks must decide how to represent mixture without forcing artificial binaries. The prestige gradient associated with diglossia shapes both human linguistic behavior and computational system outputs; when models or pipelines normalize, they tend to normalize toward MSA.

Arabic dialect writing occurs in Arabic script, Latin script (Arabizi), and mixed forms. Even within Arabic script, writers differ substantially in their orthographic choices. Several features of Arabic script are particularly relevant for understanding why orthographic variability is so pervasive in dialect writing and why preservation matters for resource creation (for a comprehensive treatment, see Habash 2010):

- **Hamza placement**: The glottal stop (*hamza*) can be written on different carriers (alif, waw, ya) or standalone, with placement rules that differ between MSA and dialect conventions. Dialect writers often omit or misplace hamza according to local pronunciation, creating systematic variation that carries phonological information.
- **Ta marbuta vs. ha**: The feminine marker *ta marbuta* () is sometimes written as *ha* () in dialectal text, reflecting pronunciation patterns. This alternation is phonologically motivated in many dialects and should not be treated as a spelling error.

- **Alif variants**: The initial alif can appear with or without hamza, and with hamza above or below ( vs.  vs. ). Dialect writers frequently use bare alif regardless of MSA rules, reflecting actual pronunciation.
- **Dialect-specific phonology in script**: Certain sounds that exist in dialects but not in MSA require creative orthographic solutions. For example, Gulf Arabic speakers may write the affricated /tš/ as *ch* in Arabizi or use the letter *kaf* () in Arabic script, creating many-to-one mappings that complicate normalization.

The CODA framework proposes a conventional orthography designed to reduce sparsity while preserving dialect identity [Habash et al., 2012], and CODA* extends these principles across multiple dialects [Habash et al., 2018]. CODA provides exactly the kind of reversible normalization layer that Guideline 3 requires; resources should maintain raw orthography as an archival layer, with CODA normalization applied as a separate, documented transformation.

## 6.2. Resource Landscape

Table 1 compares major Arabic dialect resources across dimensions relevant to LLM-assisted workflow design. This comparison highlights both the strengths of existing resources and the gaps that new projects must address.

## 6.3. Applying the Guidelines: Practical Strategies

Several guidelines have particular relevance for Arabic, and we outline both the challenges and concrete strategies for addressing them.

*Guideline 1 (Real Data Primacy):* Arabic dialect corpora should explicitly distinguish human-produced content from any synthetic augmentation. Given MSA dominance in training data, synthetic Arabic text is particularly prone to MSA drift. **Practical strategy:** Implement metadata fields at the sentence level marking provenance (human vs. synthetic), and enforce separation in storage. For shared task organizers, require participants to declare synthetic data usage and report synthetic content ratios.

*Guideline 2 (Retrieval Before Generation):* Retrieval from authenticated dialect sources reduces drift toward MSA synonyms. **Practical strategy:** Build retrieval indices over authenticated dialect corpora such as MADAR's parallel lexicon for lexicographic work, or Gumar for Gulf-specific vocabulary. Configure retrieval to prioritize dialect-specific sources and flag outputs where retrieval coverage falls below a defined threshold (e.g., fewer than 60% of generated dialect markers attested in retrieved sources).

*Guideline 3 (Orthographic Preservation):* CODA provides a principled normalization framework, but raw orthography must be maintained as the archival layer. **Practical strategy:** Store resources in a three-layer format: raw user text, CODA-normalized text, and a structured mapping log that records each normalization decision along with its type (reversible or lossy) and linguistic rationale. For the Arabic-specific features described above (hamza, ta marbuta, alif variants), the mapping log should flag instances where normalization collapses phonologically meaningful distinctions. Existing tools such as CAMeL Tools [Obeid et al., 2020] can support automated components of this pipeline.

*Guideline 4 (Dialect-Aware Prompting):* Prompts must specify the target dialect using community-preferred terminology (e.g., Egyptian, Tunisian, Gulf) and explicitly instruct against MSA substitution. **Practical strategy:** Develop prompt templates that include (a) explicit dialect specification, (b) instructions to preserve dialect-specific lexical items and morphological patterns, (c) requests to flag uncertainty rather than defaulting to MSA forms, and (d) example dialect-specific outputs to anchor generation. For instance, when generating Tunisian Arabic content, prompts should specify that negation uses *ma...sh* rather than MSA *la/lam*, and that future tense uses *besh* rather than MSA *sawfa*.

*Guideline 7 (Diversity Monitoring):* Arabic's rich dialectal variation requires monitoring along multiple dimensions simultaneously. **Practical strategy:** For each target dialect, compile a validated inventory of diagnostic markers spanning phonological spellings (e.g., Gulf *ch* for /tš/), morphological patterns (e.g., Egyptian *bi-* prefix for present tense), lexical items (e.g., Maghrebi *barsha* for "a lot"), and syntactic constructions. Track the frequency of these markers across resource versions and flag declines exceeding community-defined thresholds for manual review.

*Guideline 8 (Participatory Governance):* Arabic-speaking communities span multiple countries with distinct political contexts. **Practical strategy:** Establish advisory boards that include speakers from diverse dialect regions, balancing representation across Mashreqi and Maghrebi varieties, urban and rural speakers, and different age groups. For pan-Arabic resources, governance should involve representatives from multiple countries to avoid privileging any single national variety.

*Guideline 9 (Data Sovereignty):* Some dialect communities face political sensitivities related to regional identity or minority status. **Practical strategy:** Implement tiered access for resources containing location-identifiable content, with stricter controls for data from politically sensitive regions. Consider removing or anonymizing fine-grained geographic metadata when full access is granted.

| Resource | Coverage | Domain | Guideline Implications | LLM Workflow Role |
|---|---|---|---|---|
| AOC [Zaidan and Callison-Burch, 2011] | Multiple dialects | News comments | Retrieval grounding (G2); privacy and access control (G9) | Retrieval source for lexicographic work; dialect identification training |
| MADAR [Bouamor et al., 2018] | 25 city dialects | Travel domain | RAG grounding (G2); representativeness monitoring (G7) | Parallel lexicon for cross-dialect retrieval; generation anchoring |
| Gumar [Khalifa et al., 2016] | Gulf Arabic | Forum novels | Gulf-specific retrieval (G2); drift detection (G11) | Gulf variety benchmarking; orthographic variation archive |
| QADI [QCRI, 2020] | 18 countries | Tweets | Reproducibility (G10); documentation (G10) | Country-level dialect classifier training |
| NADI [Abdul-Mageed et al., 2021] | Country/province | Tweets | Nuanced labels (G7); privacy (G9) | Fine-grained dialect identification; evaluation benchmark |
| CODA* [Habash et al., 2012, 2018] | Cross-dialect orthography | Standardization | Reversible normalization (G3); layered representation | Normalization pipeline component; orthographic mapping reference |
| BOLT [Linguistic Data Consortium, 2021] | Egyptian Arabic | SMS/Chat | Evaluation (G5); tiered access (G9) | Egyptian variety benchmark; informal register source |
| AraP-Tweet [Zaghouani and Charfi, 2018] | 16 countries, 11 regions | Tweets | Demographic balance (G7); diversity (G8) | Multi-dialect profiling; dialect-demographic correlation |
| ADHAR [Charfi et al., 2024a] | MSA + 4 dialect groups | Tweets | Multi-dialect balance (G7); harm review (G8) | Dialect-aware hate speech detection; bias auditing |

Table 1: Arabic dialect resources with guideline implications and LLM workflow roles.

*Guideline 11 (Prestige Drift Audits):* Audits should monitor for MSA substitution and for drift toward high-visibility dialects at the expense of less-represented varieties. **Practical strategy:** At each release, sample outputs and compare dialect marker frequencies against baseline corpora. Pay particular attention to high-risk categories: function words (where MSA alternatives are common), verbal morphology (where dialect-specific conjugation patterns may be replaced by MSA forms), and discourse markers (which are highly dialect-specific). Prior work on multi-dialectal stance detection [Charfi et al., 2024b] demonstrates that annotation frameworks can effectively track cross-dialect variation, providing methodological models for drift auditing.

### 6.4. Generalizing Beyond Arabic

The Arabic case illustrates patterns that generalize to other language situations. *Diglossic languages* such as Swiss German, Haitian Creole, and Tamil exhibit similar high-low variety configurations requiring similar attention to prestige drift. *Pluricentric languages* with multiple standard varieties (German, Portuguese, Spanish) face analogous questions about which variety receives implicit priority in LLM outputs. *Languages with limited written tradition*, including many indigenous and minority languages, lack established writing systems; the orthographic preservation principle (Guideline 3) applies through documentation of transcription conventions. *Politically sensitive contexts* exist wherever communities face discrimination, persecution, or cultural appropriation risks; the data sovereignty principle (Guideline 9) generalizes accordingly.

## 7. Discussion

### 7.1. Relationship to Existing Frameworks

The proposed guidelines complement existing documentation frameworks. Datasheets for datasets

[Gebru et al., 2021] provide templates for recording corpus composition and intended use. Data statements [Bender and Friedman, 2018] emphasize speaker demographics and situational context. Model cards [Mitchell et al., 2019] document model characteristics and limitations. Our guidelines add dialect-specific requirements that existing frameworks do not fully address: orthographic layering, prestige drift monitoring, and synthetic recursion avoidance.

For community governance and data sovereignty (Guidelines 8–9), our framework aligns with the CARE Principles for Indigenous Data Governance [Carroll et al., 2020], which emphasize Collective benefit, Authority to control, Responsibility, and Ethics. While CARE was developed for indigenous communities, its principles translate well to dialect communities facing similar concerns about data extraction, misrepresentation, and loss of control over linguistic heritage [Rainie et al., 2019].

### 7.2. Implementation Challenges and Scalability

Implementing the guidelines raises practical challenges. Participatory governance (Guideline 8) requires identifying community representatives, which may be difficult for dispersed dialect populations. Diversity monitoring (Guideline 7) requires defining quantitative metrics for dialect marker presence, which presupposes linguistic analysis of diagnostic features. Long-term stewardship (Guideline 12) requires institutional commitment and sustained funding. We view these challenges as tractable rather than fundamental. The guidelines provide a framework for identifying what needs to be done; implementation pathways will vary by language, community, and institutional context.

A related concern about participatory and human-centered approaches is that they limit scalability. We argue that the guidelines support scalable practices: retrieval-augmented generation (Guideline 2) scales better than unconstrained prompting because it reduces review burden; automated drift monitoring (Guideline 11) enables early detection without exhaustive manual review; pipeline documentation (Guideline 10) enables knowledge transfer across projects. The critical constraint is that scalability must not come at the cost of distributional fidelity.

## 8. Conclusion

Large language models can support dialect resource creation when they are embedded in archival-first infrastructures, constrained by retrieval, governed by transparent documentation, and validated by dialect communities. Without such safeguards, the same systems can accelerate orthographic homogenization, prestige drift, and synthetic recursion that reduces linguistic diversity.

This paper contributes a theoretical framework, the generator-eraser paradox, and 12 community guidelines for responsible LLM-assisted dialect resource creation. The Arabic case study demonstrates application to a language situation characterized by diglossia and orthographic variability, while highlighting patterns that generalize across language families.

We invite the research community to adopt, adapt, and critique these guidelines, and to develop shared resources that make responsible LLM-assisted dialect resource creation accessible to all communities.

## 9. Limitations

This paper does not present new experiments and cannot quantify the magnitude of risks for specific models or dialects. The guidelines are intentionally conservative, prioritizing archival integrity and community governance. We have not conducted user studies with resource builders to evaluate usability. The Arabic case study, while detailed, may not capture all relevant phenomena in other diglossic or pluricentric language situations.

While we provide concrete metrics and procedures for diversity monitoring and drift audits (Guidelines 7, 11), we acknowledge that empirical validation would strengthen these recommendations. Future work should include: (1) pilot studies comparing prestige drift rates with and without dialect-aware prompting; (2) longitudinal tracking of dialect marker distributions across staged resource versions using the metrics we propose; (3) controlled comparisons of RAG-grounded versus unconstrained generation for dialect resource tasks; and (4) user studies evaluating the usability of pipeline cards and review rubrics by resource builders with varying technical backgrounds.

We also note that extending these guidelines to multimodal settings (speech, vision-language) raises additional challenges not fully addressed here, including interaction between orthographic variation and ASR/OCR errors, and the need for audio-level provenance tracking.

Future work should evaluate guideline adoption across diverse language contexts, develop shared checklists and templates as community resources, and explore how different dialect communities prioritize preservation versus standardization in applied settings.

## 10. Ethical Considerations

This work synthesizes existing research and does not collect new data. The guidelines are designed to reduce representational harm and support community sovereignty over dialect resources. We recommend tiered access, opt-out mechanisms, and participatory governance where dialect communities face safety, privacy, or cultural sensitivity risks.

We acknowledge that adopting these guidelines requires resources that may not be available to all teams, potentially creating barriers for under-resourced communities. We encourage well-resourced projects to share infrastructure, documentation templates, and lessons learned to reduce adoption barriers.

## Acknowledgments

This work was made possible by the National Priorities Research Program (NPRP) grant NPRP14C-0916-210015 from the Qatar National Research Fund (QNRF), a member of the Qatar Research, Development and Innovation Council (QRDI).

## 11. Bibliographical References

Muhammad Abdul-Mageed, Chiyu Zhang, Houda Bouamor, and Nizar Habash. NADI 2021: The second nuanced arabic dialect identification shared task. In *Proceedings of the Sixth Arabic Natural Language Processing Workshop*, 2021.

El Said Badawi. *Mustawayāt al-‘Arabiyya al-Mu‘āṣira fī Miṣr*. Dār al-Ma‘ārif, 1973.

Emily M. Bender and Batya Friedman. Data statements for natural language processing: Toward mitigating system bias and enabling better science. *Transactions of the Association for Computational Linguistics*, 6:587–604, 2018.

Emily M. Bender, Timnit Gebru, Angelina McMillan-Major, and Shmargaret Shmitchell. On the dangers of stochastic parrots: Can language models be too big? In *Proceedings of the 2021 ACM Conference on Fairness, Accountability, and Transparency*, pages 610–623, 2021.

Douglas Biber. Representativeness in corpus design. *Literary and Linguistic Computing*, 8(4): 243–257, 1993.

Steven Bird. Decolonising speech and language technology. In *Proceedings of the 28th International Conference on Computational Linguistics*, pages 3504–3519, 2020.

Su Lin Blodgett, Lisa Green, and Brendan O'Connor. Demographic dialectal variation in social media: A case study of African-American English. In *Proceedings of the 2016 Conference on Empirical Methods in Natural Language Processing*, pages 1119–1130, 2016.

Su Lin Blodgett, Solon Barocas, Hal Daumé III, and Hanna Wallach. Language (technology) is power: A critical survey of "bias" in NLP. In *Proceedings of the 58th Annual Meeting of the Association for Computational Linguistics*, pages 5454–5476, Online, 2020. Association for Computational Linguistics.

Houda Bouamor, Nizar Habash, Mohammad Salameh, Wajdi Zaghouani, Owen Rambow, Dana Abdulrahim, Ossama Obeid, Salam Khalifa, Fadhl Eryani, Alexander Erdmann, and Kemal Oflazer. The MADAR arabic dialect corpus and lexicon. In *Proceedings of the Eleventh International Conference on Language Resources and Evaluation*, 2018.

Stephanie Russo Carroll, Ibrahim Garba, Oscar L. Figueroa-Rodríguez, Jarita Holbrook, Raymond Lovett, Simeon Materechera, Mark Parsons, Kay Raseroka, Desi Rodriguez-Lonebear, Robyn Rowe, Rodrigo Sara, Jennifer D. Walker, Jane Anderson, and Maui Hudson. The CARE principles for indigenous data governance. *Data Science Journal*, 19(1):43, 2020.

J.K. Chambers. *Sociolinguistic Theory: Linguistic Variation and its Social Significance*. Wiley-Blackwell, 2013.

Anis Charfi, Wajdi Zaghouani, S. Hassan Mehdi, and Esraa Mohamed. A fine-grained annotated multi-dialectal Arabic corpus. In *Proceedings of the International Conference on Recent Advances in Natural Language Processing*, 2019.

Anis Charfi, Mabrouka Bessghaier, Raghda Akasheh, Andria Atalla, and Wajdi Zaghouani. Hate speech detection with ADHAR: A multi-dialectal hate speech corpus in Arabic. *Frontiers in Artificial Intelligence*, 7:1391472, 2024a.

Anis Charfi, Mabrouka Bessghaier, Andria Sarah Reem Atalla, Raghda Akasheh, Sara Al-Emadi, and Wajdi Zaghouani. MARASTA: A multi-dialectal Arabic cross-domain stance corpus. In *Proceedings of LREC-COLING 2024*, pages 11060–11069, 2024b.

Sasha Costanza-Chock. *Design Justice: Community-Led Practices to Build the Worlds We Need*. MIT Press, 2020.

Penelope Eckert. Three waves of variation study: The emergence of meaning in the study of sociolinguistic variation. *Annual Review of Anthropology*, 41:87–100, 2012.

Jacob Eisenstein. What to do about bad language on the internet. In *Proceedings of the 2013 Conference of the North American Chapter of the Association for Computational Linguistics: Human Language Technologies*, 2013.

Charles A. Ferguson. Diglossia. *Word*, 15(2):325–340, 1959.

Timnit Gebru, Jamie Morgenstern, Briana Vecchione, Jennifer Wortman Vaughan, Hanna Wallach, Hal Daumé III, and Kate Crawford. Datasheets for datasets. *Communications of the ACM*, 64(12):86–92, 2021.

Nizar Habash. *Introduction to Arabic Natural Language Processing*. Morgan and Claypool, 2010.

Nizar Habash, Ramy Eskander, and Abdelati Hawwari. A conventional orthography for dialectal arabic. In *Proceedings of the Eighth International Conference on Language Resources and Evaluation*, pages 711–718, 2012.

Nizar Habash, Salam Khalifa, Fadhl Eryani, Owen Rambow, Dana Abdulrahim, Alexander Erdmann, Reem Faraj, Wajdi Zaghouani, Houda Bouamor, Nasser Zalmout, Sara Hassan, Faisal Al shargi, Sakhar Alkhereyf, Basma Abdulkareem, Ramy Eskander, Mohammad Salameh, and Hind Saddiki. Unified guidelines and resources for arabic dialect orthography. In *Proceedings of the Eleventh International Conference on Language Resources and Evaluation*, 2018.

Clive Holes. *Modern Arabic: Structures, Functions, and Varieties*. Georgetown University Press, 2004.

Dirk Hovy and Anders Søgaard. Tagging performance correlates with author age. In *Proceedings of the 53rd Annual Meeting of the Association for Computational Linguistics*, 2015.

Pratik Joshi, Sebastin Santy, Amar Budhiraja, Kalika Bali, and Monojit Choudhury. The state and fate of linguistic diversity and inclusion in the NLP world. In *Proceedings of the 58th Annual Meeting of the Association for Computational Linguistics*, pages 6282–6293, 2020.

David Jurgens, Yulia Tsvetkov, and Dan Jurafsky. Incorporating dialectal variability for socially equitable language identification. In *Proceedings of the 55th Annual Meeting of the Association for Computational Linguistics*, 2017.

Salam Khalifa, Nizar Habash, Dana Abdulrahim, and Sara Hassan. Gumar: A large scale corpus of gulf arabic. In *Proceedings of the Tenth International Conference on Language Resources and Evaluation*, 2016.

Allison Koenecke, Andrew Nam, Emily Lake, Joe Nudell, Minnie Quartey, Zion Mengesha, Connor Tober, Nana Raffa, Nandini Chadha, Khushbu Das, and Sharad Goel. Racial disparities in automated speech recognition. *Proceedings of the National Academy of Sciences*, 117:7684–7689, 2020.

William Labov. *Sociolinguistic Patterns*. University of Pennsylvania Press, 1972.

Patrick Lewis, Ethan Perez, Aleksandra Piktus, Fabio Petroni, Vladimir Karpukhin, Naman Goyal, Heinrich Küttler, Mike Lewis, Wen-tau Yih, Tim Rocktäschel, Sebastian Riedel, and Douwe Kiela. Retrieval-augmented generation for knowledge-intensive NLP tasks. In *Advances in Neural Information Processing Systems*, 2020.

Linguistic Data Consortium. Bolt egyptian arabic treebank: Sms and chat. https://catalog.ldc.upenn.edu/LDC2021T17, 2021. Accessed 2026-02-28.

Tony McEnery and Andrew Hardie. *Corpus Linguistics: Method, Theory and Practice*. Cambridge University Press, 2012.

Margaret Mitchell, Simone Wu, Andrew Zaldivar, Parker Barnes, Lucy Vasserman, Ben Hutchinson, Elena Spitzer, Inioluwa Deborah Raji, and Timnit Gebru. Model cards for model reporting. In *Proceedings of the Conference on Fairness, Accountability, and Transparency*, pages 220–229, 2019.

Ossama Obeid, Go Inoue, Bashar Alhafni, Salam Khalifa, and Nizar Habash. CAMeL tools: An open source python toolkit for arabic natural language processing. In *Proceedings of the Twelfth International Conference on Language Resources and Evaluation*, 2020.

QCRI. QADI: QCRI arabic dialects identification corpus. https://alt.qcri.org/resources/qadi/, 2020. Accessed 2026-02-28.

Stephanie Carroll Rainie, Tahu Kukutai, Maggie Walter, Oscar Luis Figueroa-Rodríguez, Jennifer Walker, and Per Axelsson. Issues in open data: Indigenous data sovereignty. In Tim Davies, Stephen B. Walker, Mor Rubinstein, and Fernando Perini, editors, *The State of Open Data: Histories and Horizons*, pages 300–319. African Minds and International Development Research Centre, Cape Town and Ottawa, 2019.

Maarten Sap, Dallas Card, Saadia Gabriel, Yejin Choi, and Noah A. Smith. The risk of racial bias in hate speech detection. In *Proceedings of the 57th Annual Meeting of the Association for Computational Linguistics*, pages 1668–1678, 2019.

Ilia Shumailov, Zakhar Shumaylov, Yiren Zhao, Nicolas Papernot, Ross Anderson, and Yarin Gal. Ai models collapse when trained on recursively generated data. *Nature*, 631:755–759, 2024.

Samson Tan, Shafiq Joty, Min-Yen Kan, and Richard Socher. It's morphin' time! combating linguistic discrimination with inflectional perturbations. In *Proceedings of ACL*, 2020.

Kees Versteegh. *The Arabic Language*. Edinburgh University Press, 2014.

Wajdi Zaghouani. Critical survey of the freely available Arabic corpora. In *Proceedings of the Workshop on Free/Open-Source Arabic Corpora and Processing Tools, LREC 2014*, 2014.

Wajdi Zaghouani and Anis Charfi. Arap-tweet: A large multi-dialect twitter corpus for gender, age and language variety identification. In *Proceedings of the Eleventh International Conference on Language Resources and Evaluation*, 2018.

Wajdi Zaghouani, Hamdy Mubarak, and Md Rafiul Biswas. So hateful! building a multi-label hate speech annotated Arabic dataset. In *Proceedings of LREC-COLING 2024*, pages 15044–15055, 2024.

Omar F. Zaidan and Chris Callison-Burch. The arabic online commentary dataset: An annotated dataset of informal arabic with high dialectal content. In *Proceedings of the 49th Annual Meeting of the Association for Computational Linguistics: Human Language Technologies*, pages 37–41, 2011.

Marcos Zampieri, Shervin Malmasi, Preslav Nakov, et al. Findings of the VarDial evaluation campaign. In *Proceedings of the Fourth Workshop on NLP for Similar Languages, Varieties and Dialects*, 2017.